\pdfoutput=1

\documentclass[11pt]{article}
\usepackage{graphicx}
\usepackage{algorithmic}
\usepackage{algorithm}
\usepackage{arydshln}
\usepackage{acl}

\usepackage{times}
\usepackage{latexsym}

\usepackage[T1]{fontenc}

\usepackage[utf8]{inputenc}

\usepackage{microtype}

\title{Can Offline Reinforcement Learning Help Natural Language Understanding?}

 \author{Ziqi Zhang\thanks{\,\, Equal contribution.}, Yile Wang\footnotemark[1], Yue Zhang \and Donglin Wang\\
         \texttt{stevezhangz@163.com}\\
 \texttt{\{wangyile,zhangyue,wangdonglin\}@westlake.edu.cn} \\ 
            School of Engineering, Westlake University\\}

\begin{document}
\maketitle
\begin{abstract}
Pre-training has been a useful method for learning implicit transferable knowledge and it shows the benefit of offering complementary features across different modalities. Recent work mainly focuses on the modalities such as image and text, for example, studies show that visual features learned from images can help visual-grounded language understanding. In this paper, we consider investigating the potential connection between offline reinforcement learning (RL) and language modeling (LM). Intuitively, RL and LM are similar in predicting the next states based on the current and previous states, which rely on both local and long-range dependency across states. To validate such an assumption, we pre-trained different offline RL tasks using Transformer and then evaluate these models on various language-related tasks. Experimental results show that our RL pre-trained models can give close performance compared with the models using the LM training objective, showing that there exist common useful features across these two modalities. To further explore the potential relationship, we investigate some factors such as Markov property and the sequential nature of RL trajectory.

\end{abstract}

\section{Introduction}

\begin{figure*}[h]
    \centering
    \includegraphics[scale=0.7]{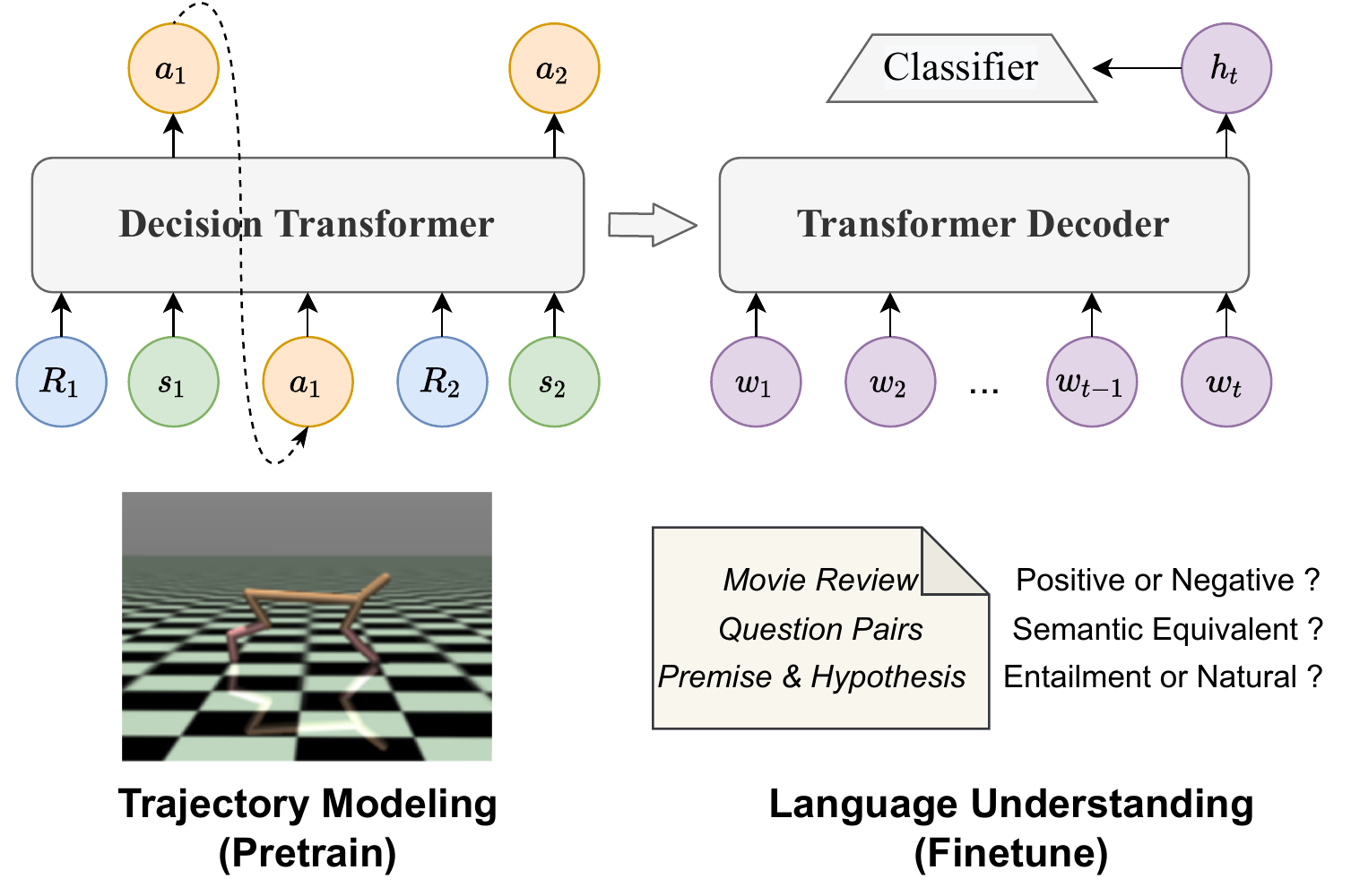}
    \caption{Adapting reinforcement learning pre-trained model to natural language understanding tasks.}
    \label{figure:pre-train-finetune}
\end{figure*}

\noindent
Transformer~\cite{attentionisalluneed} based architecture have been widely used in natural language processing (NLP), showing strong performance in various tasks including natural language understanding (NLU)~\cite{wang2018glue}, question answering~\cite{rajpurkar2016squad,rajpurkar2018know}, and summarization~\cite{hermann2015teaching,hasan2021xl}. Recent studies show that Transformer can also be used in computer vision~\cite{ViT,swin}, speech~\cite{speechtransformer}, and biological application~\cite{biotransformer,transformerprotein}.
The unifying of model architecture makes the gap between different data modalities closer. For example, many multi-modal representations based on Transformer have been proposed~\cite{transformervideoretrivel,NEURIPS2020_ff0abbcc}, and there are also works show that visual features can help language representations~\cite{tan-bansal-2020-vokenization,zhang2022mcse}.

Chen et al. proposed Decision Transformer(DT) via a sequence modeling method to tackle offline reinforcement learning tasks~\cite{DT}. Specifically, DT treats the return, state and action at each time step as sequential data $R_1, s_1, a_1, ..., R_{i}, s_{i}, a_{i}$, which is similar to the GPT2~\cite{GPT2}, a language model that builds the left-to-right token dependency through deep uni-directional self-attention neural network.


Pre-training on corpora such as Wikipedia and book-corpus has shown useful for improving the performance of downstream language-related tasks, due to the capability of learning useful features from large-scale data~\cite{elmo,BERT}. Unlike languages, RL pre-training is a challenge due to the lack of datasets~\cite{cwho}. Intuitively, the trajectories in offline reinforcement learning share some characteristics with natural language. For example, both offline reinforcement learning and language datasets share the sequential nature of states or tokens. However, the relationship between these two different modal scenarios still remains unclear and the potential connection has not been fully exploited. 



Benefiting from the unifying of model architecture (e.g., DT and GPT2), we explore the underlying similarities between offline RL tasks and languages. In particular, we pre-train DT on RL tasks and then evaluate the performance of these models on various NLU tasks to examine whether offline RL pre-training can help language understanding. Our results show that offline RL tasks can generally improve model performance on NLP tasks, and even show some competitive and better results than models using LM training objective.

In summary, our contribution could be concluded as follows:

$\bullet$ We first explored the relationship between RL and natural language tasks by using the pre-train and finetune methodology. Evaluating DT trained on RL tasks and transferring it to six NLU tasks, we find that our pre-trained DT surpasses the un-pre-trained Transformer model and gives competitive or even better results than LM pre-trained GPT on the wikitext-103 dataset.

$\bullet$ We further analyze the potential common features between language and RL, finding that (1) Markov property leads to some improvement but it's not a decisive factor; (2) Sequential nature of RL trajectory can be served as a useful feature for language understanding.

\section{Related Work}
{\bf Transformer.}  Transformer is based on a kind of deep self-attention network and was first proposed by ~\citet{attentionisalluneed} for neural machine translation. Based on its powerful representation capability, transformer variants have been widely used for learning general features in different domains such as natural language processing, computer vision, speech, and reinforcement learning, typical models include GPT~\cite{GPT2}, BERT~\cite{BERT}, ViT~\cite{ViT}, wave2vec~\cite{wav2vec}, and Decision Transformer~\cite{DT}. The unified Transformer backbone narrows the difference between different modalities and prompts the research of cross-modal research. \\
\noindent
{\bf Cross-modal Knowledge Transfer.}  Recent work shows that language tasks benefit from other domain tasks such as visual pre-training. For example, Tan and Bansal improve language understanding by contextually mapping language tokens to their related images~\cite{tan2020vokenization}. Chen and Li proposed a large-scale pre-trained model for image-text representation learning~\cite{chen2020uniter}. Radford and Alec et al. proposed learning image representation by matching the image to its caption~\cite{radford2021learning}. 
Huo et al. proposed a large-scale multi-modal contrastive learning pre-training model~\cite{huo2021wenlan}. 

\noindent
{\bf Relationship between RL and Natural Language.} Due to the lack of large-scale off-the-shelf datasets, offline RL is difficult and challenging to transfer among different environments. To leverage different potential useful resources, ~\citet{cwho} pre-train the DT by language modeling on natural language datasets such as wikitext-103, showing surprising results that languages-based pre-training may help improve the offline RL. However, they did not further exploit the underlying similarity between these two modalities. Our work can be complementary, where we try different models by RL and investigate whether and how they can help general language understanding, trying to reveal the relationship such as Markov property and sequential nature of both language and RL. 
\section{Methodology}
\noindent
{\bf Transformer for Language Modeling.} Language modeling is a fundamental task in NLP which aims to autoregressively predict token $x_t$ based on the input tokens $x_1, x_2, \cdots, x_{t-1}$.
\begin{equation}\label{eq1}
   P(x)=\prod \limits_{t=0}^n {P}(x_t|x_1,x_2, \cdots, x_{t-1})
\end{equation}

GPT~\cite{GPT2} models each token $x_i$ by multi-layer Transformer decoder, which interacted with all previous tokens by uni-directional attention mechanism, and predicts the token $x_t$ by the hidden states of token $x_{t-1}$:

\begin{equation}
h_1,h_2,...,h_t= {\rm Embeddings}(x_1,x_2,...,x_t) \\
\end{equation}
\begin{equation}
h_t= {\rm Transformer\_Decoder}(h_0,...,h_{t-1}) \\
\end{equation}
\begin{equation}
P(x_t)={\rm Softmax}(Wh_{t-1})\\
\end{equation}
\label{eq2}

where Embeddings denote the token and position embedding table, $W$ is a trainable language modeling head.
Pre-training on large text corpus, GPT can be useful for learning transferable knowledge of the language and improving the performance of the downstream task by plugging in specific classification head $W_{cls}$:
\begin{equation}
P(x_1,x_2,...,x_t)={\rm Softmax}(W_{cls}h_{t}) 
\end{equation}

\noindent
{\bf Offline Reinforcement Learning and Decision Transformer.}
We consider a general Markov Decision Process (MDP) as $(\mathcal{S}, \mathcal{A}, \mathcal{P}, \mathcal{R})$. In a MDP tuple, $\mathcal{S}$ represents state space,  $\mathcal{A}$ represents action space, $\mathcal{P}(s_{t+1}|s_{t},a_{t})$ denotes the transition function at time step $t$, $ \mathcal{R}$ is the reward space. In particular, $R(s_t,a_t) \in \mathcal{R} $ represents the reward function which is used to get the reward $r_t$. Hence, the return-to-go $R$ could be described as $R_T=\sum_{t=t'}^{t=T} r_{t}$. In addition, RL aims to find the optimal policy $\pi^*$ and gain the maximum return. According to whether directly interact with the environment, RL could be divided into offline RL and online RL. Different from online RL, offline RL makes full use of static dataset to train a model without interaction with the environment.

Decision Transformer (DT)~\cite{DT} frames offline RL tasks as sequential modeling tasks to autoregressively predict action $a_t$ based on the historical information $\{R_{t-K}, s_{t-K}, a_{t-K}, R_{t-K+1}, s_{t-K+1}, a_{t-K+1}, .$
$.., R_{t}, s_{t}\}$. Specifically, DT used pruned version of Transformer decoder as its backbone, additionally using linear layers to embed elements of trajectory $\{R_{t-K}, s_{t-K}, a_{t-K}, R_{t-K+1}, s_{t-K+1}, a_{t-K+1}, .$
$.., R_{t}, s_{t}\}$ into a unified representation space, together with time step embedding to represent the position feature. The training objective follows a way similar to language modeling to infer actions. During training, DT can be optimized by minimizing the mean squared error between the generated action and the ground truth.

\noindent\textbf{pre-train and Finetune.} As shown in Figure.~\ref{figure:pre-train-finetune}, to investigate the relationship between language and reinforcement learning, we use Transformer architecture as our backbone and pre-train it on different RL tasks, then compare these pre-trained models with language modeling(LM) pre-trained Transformer models in various NLU tasks. In particular, we first pre-trained DT in 11 tasks from d4rl~\footnote{\url{https://github.com/rail-berkeley/d4rl}} benchmark, then transfer these models for language-related tasks, including sentiment analysis, paraphrase matching, and natural language inference.



\section{Experiments}
\subsection{Dataset}

\begin{table*}[h]
  \centering
  \begin{tabular}{lccc}
    \hline
    {\bf Dataset Name}     &  {\bf Domain}    & {\bf Decision Type}   & {\bf Markov}\\
    \hline
    hopper-expert-v2 & Gym-Mujoco  & Policy   & High  \\
    walker2d-expert-v2    & Gym-Mujoco & Policy & High      \\
    halfcheetah-expert-v2     & Gym-Mujoco      & Policy & High  \\
            hopper-medium-replay-v2 & Gym-Mujoco  & Policy  & Medium    \\
    walker2d-medium-replay-v2     & Gym-Mujoco & Policy  & Medium      \\
    halfcheetah-medium-replay-v2     & Gym-Mujoco       & Policy & Medium \\
        hopper-random-v2 & Gym-Mujoco & Policy  & Weak    \\
    walker2d-random-v2     & Gym-Mujoco & Policy  & Weak      \\
    halfcheetah-random-v2     & Gym-Mujoco      & Policy & Weak  \\

        maze2d-large     & Maze2d  & Controller  & No    \\
    antmaze-large-diverse     & Antmaze       & Controller & No  \\
    \hline
  \end{tabular}
    \caption{Our offline reinforcement learning tasks from d4rl. }
  \label{ourd4rltasks}
\end{table*}
\noindent
{\bf Offline Reinforcement Learning Dataset.} D4rl benchmark is composed of various kinds of offline RL datasets from different domains~\cite{d4rl}.  As shown in  Table.\ref{ourd4rltasks}, we mainly pre-train DT on three domains including Maze2d, Antmaze, and Gym-Mujoco. Maze2d and Antmaze are collected by the interaction of a human-designed controller with the environment, which does not rely on the Markov property. In terms of Gym-Mujoco, it is collected through the interaction between three level policies and the environment. In particular, the offline RL datasets of Gym-Mujoco (Halfcheetah, Walker2d, Hopper), can be divided into three types according to the level of policy and Markov property:

{$\bullet$ {\textit{Expert}}.} This type is collected by the interaction of expert policy with the environment. The Markov property of these datasets is high.

{$\bullet$ {\textit{Medium Replay}}.} It includes all samples in the replay buffer during the training process of a policy till it reaches the medium level ("Medium" means first training a SAC~\cite{SAC} agent with early stopping, and using the early stopped policy to collect the dataset).

{ $\bullet$ {\textit{Random}}.} It is a million steps dataset generated by the interaction of random policy with the environment. The Markov property of these datasets is weak.

\noindent
{\bf Natural Language Understanding Datasets.} We select six NLU tasks from GLUE~\cite{wang2018glue} benchmark for evaluation. Those tasks can be grouped into three categories: (1)  single-sentence classification: SST2; (2) similarity and paraphrase matching: MRPC, QQP; (3) natural language inference: WNLI, QNLI, RTE.

{$\bullet$ {\textit{SST2}}.} The Stanford Sentiment Treebank can be used for text classification tasks, by predicting the sentiment of a sentence among positive or negative labels.

{$\bullet$ {\textit{MRPC}}.} The Microsoft Research Paraphrase Corpus is used for judging whether two sentences are semantically equivalent.

{$\bullet$ {\textit{QQP}}.} The Quora Question Pairs dataset is used to semantically analyze whether two questions have the same meaning.

{$\bullet$ {\textit{QNLI}}.} The Stanford Question Answering Dataset is used to determine whether the context sentence contains the answer to the question.  

{$\bullet$ {\textit{WNLI}}.} This task comes from the Winograd Schema Challenge, which aims to predict if the sentence with the pronoun substituted is entailed by the original sentence.

{$\bullet$ {\textit{RTE}}.} The Recognizing Textual Entailment comes from textual entailment challenges, which aim to judge whether the premise entails the hypothesis or not

\subsection{Baselines}
We compare different models for NLU tasks:

\noindent{\bf RL-GPT.} Decision transformer (DT) is proposed using sequence modeling method to tackle offline RL tasks. Our DT has 256 embedding dim, 8 heads, and 3 layers. In particular, it adds three linear embedding layers on normal GPT architecture to unify the state space $\mathcal{S}$,  action space $\mathcal{A}$, and reward space $\mathcal{R}$. After pre-training on RL tasks, we extract the GPT sub-module and named it RL-GPT for NLU tasks finetuning.


\noindent{\bf LM-GPT.} To compare our RL-GPT with model pre-trained using language modeling objective on normal text. We pre-train a Transformer decoder with the same parameter setting of RL-GPT on wikitext-103 dataset, which we named as LM-GPT. 
   
\noindent{\bf Random-GPT.} To verify the effectiveness of RL and LM pre-training, we also finetune a randomly initialized Transformer decoder on different downstream tasks. We call this baseline Random-GPT.
   
\begin{table*}[h]
    \small
  \centering
  \begin{tabular}{lccccccc}
    \hline
     {\bf Model} & {\bf SST2}& {\bf MRPC} & {\bf QQP}  & {\bf QNLI} & {\bf WNLI}& {\bf RTE}  &{\bf Avg.} \\
    \hline
    
    Random-GPT  	&  \underline{79.91±0.85} & 69.9±0.39& 76.95±0.15 & 61.33±0.31	& 47.61±5.8& 48.88±1.13	&64.10\\
    \hdashline 
    hopper-expert &79.06±0.26&	69.46±0.78	&78.41±0.1&		\underline{61.35±0.18}&	53.8±1.38&  51.12±2.99 &65.53\\
    hopper-medium-replay &78.99±0.78&	69.22±0.43&\underline{78.44±0.12}	&	61.15±0.28&	49.3±2.95&  51.26±0.6	&64.73\\
    hopper-random& 79.7±1.14&	69.07±0.42&	77.35±0.05 	&	60.98±0.27	&51.55±3.52& 48.45±0.53&64.51\\
    walker2d-expert& 77.78±0.61 & 	69.85±0.76& 77.74±0.14 &		60.52±0.27& 	52.11±5.12& 49.68±2.53&64.61\\
    walker2d-medium-replay&79.08±0.41& 70.05±0.68&78.33±0.1 & 61.04±0.2&52.96±6.34&50.76±2.16&65.37\\
    walker2d-random &73.51±0.95&68.68±0.18& 76.93±0.07 &59.34±0.57& 51.83±3.01&52.27±4.22&63.76\\
    halfcheetah-expert &77.98±0.58&	69.61±0.51	&77.5±0.09 &	61±0.39&	49.86±1.69&52.27±2.15&64.70\\
    halfcheetah-medium-replay &	79.59±0.63&\underline{70.69±0.95} &77.73±0.22	&60.54±0.38&	50.7±4.23& 51.77±2.1&65.17\\
    halfcheetah-random &74.04±0.95&	68.92±0.33&76.92±0.08&59±0.47	&50.99±4.66 & 52.42±1.77	&63.72\\
    antmaze-large-diverse &76.86±0.59&	70±0.74&	77.37±0.16  &	60.37±0.33 & \underline{54.65±5.94}& 53.29±1.42&65.42\\
    maze2d-large& 77.5±0.6& 	67.91±0.37& 	77.67±0.18& 	60.42±0.33	& {\bf 54.93±3.98}& \underline{52.78±2.77}&65.20\\
    \hdashline 
    LM-GPT &{\bf 82.43±0.64} &{\bf 71.81±0.78}&	{\bf 78.62±0.05} &{\bf 63.22±0.38}&	51.55±4.85& {\bf 56.97±0.42}&67.43\\
    \hline
  \end{tabular}
      \caption{Comparison for results on six NLU tasks. The second best data has been underlined}
  \label{NluResults}
\end{table*}

\subsection{Settings}
\noindent
{\bf RL-pre-train.} Following \citet{DT} and \citet{cwho}, we train DT on offline RL tasks, including Gym-Mujoco, Antmaze, and Maze2d. Specifically, we set learning rate 1e-4, batch size 64, max length 20, warmup steps 10000, and train total 40 epoches.


\noindent
{\bf LM-pre-train.} We train our LM-GPT through language modeling on wikitext-103~\cite{merity2016pointer}. We set learning rate 5e-5,  batch size 4, total training steps 140k and used Adam~\cite{kingma2014adam} for optimization. 

\noindent
{\bf Finetuning on NLU tasks.}
We evaluate RL-GPT, LM-GPT, and Random-GPT  on SST2, MRPC, QQP, QNLI, WNLI and RTE. Specifically, we set max length 128, batch size 128, learning rate 2e-5.

\subsection{Results}
Our experiment results are shown in Table.~\ref{NluResults}. RL-GPT pre-trained on antmaze-large-diverse, maze2d-large, halfcheetah-medium-replay, and walker2d-medium-replay achieved competitive performance that its average accuracy is 65. In addition, the average accuracy of LM-GPT is 67.43, which is the best on average among our six NLU tasks.   In addition, the average accuracy of Random-GPT is 64.10. 

In particular, the average accuracy of 8 RL-GPT models exceeded Random-GPT, which proved that pre-training on the offline reinforcement learning dataset can give improvement to NLU tasks. 

In addition, for each task, the results are slightly different where we find:

(1) For the SST2 task, the Random-GPT gives the second-best results and the RL-GPT models perform generally poorly. The reason can be that the sentiment of a sentence does not rely much on the sequential text compared with the sentiment polarity words, so RL pre-training does not help.

(2) For sentence-pair matching tasks (MRPC and QQP), most of the results are better than Random-GPT (about 1-2 point improvement).

(3) In RTE task, compared to Random-GPT, pre-training on offline RL datasets can bring about 1 to 5 points of improvement. Specifically, for WNLI,  RL-GPT tasks will give an obvious enhancement. What's more, compared to Random-GPT, only pre-train RL-GPT on hopper-expert won't give obvious enhancement to RTE, and pre-train RL-GPT on other offline RL tasks can bring about 2 to 7 points of improvement. Besides, compared to pre-training on wikitext-103, RL-GPT pre-trained on walker2d-expert, walker2d-medium-replay, antmaze-large-diverse, and maze2d-large have more competitive performance. 

Overall, pre-training on RL tasks can help NLU tasks broadly, especially for the sentence-pair matching and NLI tasks, where sequential information is crucial for semantic understanding and inference.

\begin{table*}[h]
\small
\centering
\begin{tabular}{lccccccc}
\hline
     {\bf Model} & {\bf SST2} & {\bf MRPC} & {\bf QQP} & {\bf QNLI} & {\bf WNLI} & {\bf RTE} &{\bf Avg.} \\  
    \hline
    walker2d-mr& 79.17±0.46&70.34±0.35&78.47±0.06&61.32±0.32&51.83±4.75&51.12±2.88&65.38\\
    \hdashline 
    inner-inter-shuffle  &80.64±0.73&69.75±0.2&77.22±0.12&61.75±0.33&47.61±4.99&50.25±1.54&64.54\\
    inner-shuffle  &80.28±0.75&70.2±0.37&78.66±0.06&62.16±0.42&51.27±5.75&52.64±1.15&65.87\\
    inter-shuffle &79.45±0.58	&68.97±0.67	&78.63±0.1	&61.82±0.72	&54.08±3.29	&52.49±1.22	&65.91 \\
    \hline
  \end{tabular}
  \caption{Results of models pre-trained on shuffled walker2d-medium-replay dataset.}
       \label{shuffle-walker2d}
\end{table*}
\begin{table*}[h]
\small
\centering
\begin{tabular}{lccccccc}
\hline

     {\bf Model} & {\bf SST2} & {\bf MRPC} & {\bf QQP} & {\bf QNLI} & {\bf WNLI} & {\bf RTE} &{\bf Avg.} \\  
    \hline
    antmaze-large-diverse & 78.49±0.35 & 69.95±0.5	& 78.5±0.17&60.91±0.37&53.52±2.52&53.5±1.52&	65.81\\
    \hdashline 
    inner-inter-shuffle &77.82±0.41&70.29±0.61&77.5±0.2&61.29±0.43&55.77±3.74&53.36±1.32&66.01\\
    inner-shuffle  &80.16±0.34&	69.12±0.82&	78.65±0.06	&61.52±0.27	 &52.39±5.45&	52.49±1.4&	65.72\\
    inter-shuffle &79.43±0.66&69.07±0.65&78.6±0.14&61.44±0.41&51.55±1.91&52.64±1.01	&65.46 \\
    \hline
  \end{tabular}
    \caption{Results of models pre-trained on shuffled antmaze-large-diverse  dataset.}
          \label{shuffle-antmaze}
\end{table*}

\section{Discussion}

As shown in Table.\ref{NluResults}, most (8 of 11) of our RL-GPT models' average accuracy surpass Random-GPT models' on we selected NLU tasks, and we consider two factors that may affect the model performance that the first is the Markov property, and another is the sequential nature of tokens. To validate this, we set up a series of experiments to investigate and analyze.

\noindent
{\bf The Markov property of offline RL tasks.}  RL is generally considered a Markov decision process, and natural language also has Markov property. Intuitively, the improvement given by pre-trained on offline RL tasks is because both offline RL dataset and NLP dataset have Markov property. However, this inference is not so rigorous, because the domain of Maze2d and Antmaze are both relied on human-designed controllers instead of policy models with Markov property, but RL-GPT models pre-trained on those domains give the best average accuracy among all kinds of RL-GPT models. In addition, according to the way of data collection, the Markov property of the expert dataset is stronger than the medium-replay dataset, but we can list many results that pre-training on medium-replay datasets gives stronger performance compared to pre-training on the expert dataset.

$\bullet$  Pre-training RL-GPT on walker2d and finetune on RTE, MRPC, SST2, QNLI and QQP.

$\bullet$  Pre-training RL-GPT on halfcheetah and finetune on MRPC, SST2, and WNLI.

Therefore, we can induce that the Markov property of offline RL datasets isn't the necessary factor that determines RL-GPT performance on NLU tasks.

Although the Markov property isn't the deceiving factor influencing the performance of RL-GPT, pre-training RL-GPT on some RL datasets collected by random policy can not significantly improve it. For example, compared to the random datasets, expert and medium-replay datasets have stronger Markov properties. However, in most NLU tasks, the improvement of RL-GPT brought by pre-training on expert or medium-replay datasets is often more obvious than on random datasets. For instance:

$\bullet$ Pre-training RL-GPT on halfcheetah and finetune on MRPC, SST2, QNLI, QQP. 

$\bullet$ Pre-training RL-GPT on walker2d and finetune on MRPC, SST2, QNLI, WNLI, QQP.

$\bullet$ Pre-training RL-GPT on hopper and finetune on RTE, MRPC, QNLI, QPP. 

All those experiments show that RL-GPT models pre-trained on expert or medium-replay datasets perform better than pre-trained on the random datasets, which supports the view that although the Markov property isn't the decisive factor affecting the model performance, too random datasets can not bring obvious improvement to the NLU tasks.

\begin{figure}[htbp]
    \centering
    \includegraphics[scale=0.27]{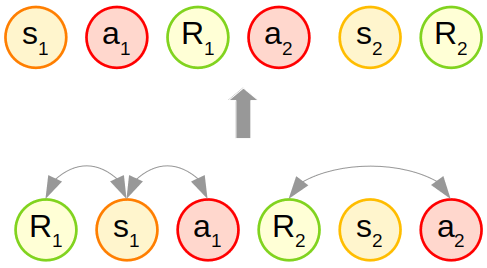}
    \caption{Illustration of our inter shuffling settings.}
        \label{inter}
\end{figure}

\begin{figure}[htbp]
    \centering
    \includegraphics[scale=0.27]{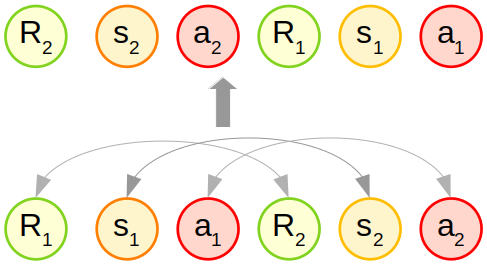}
    \caption{Illustration of our inner shuffling settings.}
    \label{inner}
\end{figure}

\noindent
{\bf Influence of the sequential nature of tokens.} In the above analysis, we have found that the Markov property is not the decisive factor in the performance of RL-GPT, so there must be other factors affecting the performance of RL-GPT. We consider the order and sequential nature of datasets as a potential factor, and to validate this, We set up three methods to shuffle the input tokens. Our methods can be described by Figure.\ref{inter} and Figure.\ref{inner}. Among them,  Figure.\ref{inter} focuses on shuffling the position of different kinds of tokens without changing the positions of one specific kind of tokens, and we named this method inter-shuffle, on the contrary, Figure.\ref{inner} focuses on shuffling the position of one specific kind of tokens without changing the relative positions among different kinds of tokens, we named this method as inner-shuffle. In particular, we also combine inner-shuffle and inter-shuffle to process data, which is inner-inter-shuffle. We run those experiments on walker2d-medium-replay and antmaze-large, which can give GPT obvious enhancement on our NLU tasks.

We pre-train RL-GPT on the shuffled datasets, and according to the average accuracy of the pre-trained RL-GPT on our NLU tasks(Table.\ref{shuffle-antmaze} and Table.\ref{shuffle-walker2d}), we got the following conclusions:

(1) When pre-training on walker2d-medium-replay, only completely disrupting the order of tokens (using the inner-inter-shuffle dataset) can decrease the performance of RL-GPT.

(2) When pre-training on antmaze-large-diverse, the completely disrupted datasets can not significantly reduce the performance of RL-GPT on our NLU tasks, and processed antmaze-large-diverse with inter-shuffle or inner-shuffle can reduce the performance of RL-GPT.

In summary, the sequential nature of tokens can influence the performance of RL-GPT on NLU tasks. However, the relationship between the sequential nature of tokens and the performance of the RL-GPT still needs to be further explored and studied.

\section{Conclusion} In this research, we finetune RL-GPT, LM-GPT, and Random-GPT on our NLU tasks(SST2, MRPC, QQP, QNLI, WNLI, RTE), and the results show that most of our RL tasks give enhancement to the results of RL-GPT evaluation on our NLU tasks, especially gives an obvious enhancement to RTE, WNLI, and QQP.

In addition, we explore which factor may affect the model performance. The first is the Markov property, and the second is the sequential nature. Specifically, by comparing our 11 RL-GPT models we find Markov property may not be a decisive factor in the performance of RL-GPT. As for the sequential nature of tokens, we use three methods includes inner-shuffle, inter-shuffle, and inter-inner-shuffle to process RL tasks and investigate the influence of the sequential nature of tokens on NLU tasks. Experimental results show that the sequential nature of token can influence the model performance.

To sum up, RL tasks can help RL-GPT train on NLU tasks, and the performance of RL-GPT is influenced by many factors, including but not limited to Markov property, and the sequential of tokens. Among them, Markov property isn't the decisive factor that influence RL-GPT performance. As for the sequential nature of tokens, pre-trained on the shuffled dataset can influence model performance. However, the relationship between the sequential nature and model performance still needs to be further explored.

\bibliography{anthology}
\bibliographystyle{acl_natbib}




\end{document}